\documentclass{article}

\usepackage{arxiv}
\usepackage{authblk}
\usepackage[utf8]{inputenc} % allow utf-8 input
\usepackage[T1]{fontenc}    % use 8-bit T1 fonts
\usepackage{hyperref}       % hyperlinks
\usepackage{url}            % simple URL typesetting
\usepackage{booktabs}       % professional-quality tables
\usepackage{amsfonts}       % blackboard math symbols
\usepackage{nicefrac}       % compact symbols for 1/2, etc.
\usepackage{microtype}      % microtypography
\usepackage{lipsum}
\usepackage{graphicx}
\usepackage{amssymb}
\usepackage{amsmath}
\usepackage{algorithmic}
\usepackage{algorithm}
\usepackage{textcomp}
\usepackage{stfloats}
\usepackage{verbatim}
\usepackage{multirow}
\usepackage{subfigure}
\usepackage{array}
\usepackage{longtable}
\graphicspath{ {./images/} }

\title{Trunk-branch Contrastive Network with Multi-view Deformable Aggregation for Multi-view Action Recognition}

\author[1,2]{Yingyuan Yang}
\author[3,*]{Guoyuan Liang}
\author[3]{Can Wang}
\author[4]{Xiaojun Wu}

\affil[1]{Shenzhen Institutes of Advanced Technology, Chinese Academy of Sciences, Shenzhen 518055, China.}
\affil[2]{University of Chinese Academy of Sciences, Beijing 10049, China.}
\affil[3]{Guangdong Provincial Key Laboratory of Robotics and Intelligent System, Shenzhen Institutes of Advanced Technology, Chinese Academy of Sciences, Shenzhen 518055, China.}
\affil[4]{School of Mechanical Engineering and Automation, Harbin Institute of Technology at Shenzhen, Shenzhen 518055, China.}
\begin{document}
\maketitle
\begin{abstract}
Multi-view action recognition aims to identify actions in a given multi-view scene. Traditional studies initially extracted refined features from each view, followed by implemented paired interaction and integration, but they potentially overlooked the critical local features in each view. When observing objects from multiple perspectives, individuals typically form a comprehensive impression and subsequently fill in specific details. Drawing inspiration from this cognitive process, we propose a novel trunk-branch contrastive network (TBCNet) for RGB-based multi-view action recognition. Distinctively, TBCNet first obtains fused features in the trunk block and then implicitly supplements vital details provided by the branch block via contrastive learning, generating a more informative and comprehensive action representation. Within this framework, we construct two core components: the multi-view deformable aggregation (MVDA) and the trunk-branch contrastive learning. MVDA employed in the trunk block effectively facilitates multi-view feature fusion and adaptive cross-view spatio-temporal correlation, where a global aggregation module (GAM) is utilized to emphasize significant spatial information and a composite relative position bias (CRPB) is designed to capture the intra- and cross-view relative positions. Moreover, a trunk-branch contrastive loss is constructed between aggregated features and refined details from each view. By incorporating two distinct weights for positive and negative samples, a weighted trunk-branch contrastive loss is proposed to extract valuable information and emphasize subtle inter-class differences. The effectiveness of TBCNet is verified by extensive experiments on four datasets including NTU-RGB+D 60, NTU-RGB+D 120, PKU-MMD, and N-UCLA dataset. Compared to other RGB-based methods, our approach achieves state-of-the-art performance in cross-subject and cross-setting protocols.
\end{abstract}

% keywords can be removed

\section{Introduction} 

Human action recognition (HAR) has emerged as an active and significant research domain in computer vision. The advancement in this field has markedly augmented the level of intelligence across various sectors like public safety and entertainment \cite{das2023viewclr}.
Given the massive increase in the number of surveillance and traffic monitoring cameras, the HAR task has broadened to encompass multi-view scenarios.
Furthermore, as the majority of cameras in real-world multi-view scenes only offer RGB video, our works predominantly concentrate on the RGB modality for multi-view action recognition.

In contrast to action recognition from a single, static viewpoint, multi-view action recognition necessitates the effective amalgamation of action features captured by multiple cameras. 
However, due to differences in action features and visible appearances observed from various viewpoints, an inadequate integration may lead to poor classification results. 
Numerous methods \cite{wang2018dividing, ULLAH2021321} had been developed to extract robot and inter-reliant features.
DANet \cite{wang2018dividing} employed conditional random field to pass message among view-specific features from different views and weighted multi-view classification results using the prediction probability of each view. 
%Following the extraction of self-reliant patterns from the views, the conflux LSTMs network \cite{ULLAH2021321} proceeded to conclude interactions and inter-reliant patterns by pairwise dot, thereby highlighting the similarities from various viewpoints.
However, these methods present challenges in constructing a comprehensive correlation among multiple views, and they may disregard local effective features in low-accuracy views.
Recent methods \cite{liu2023dual,siddiqui2024dvanet} utilized disentangled representation learning to accomplish the separation and interaction between the action and view, capturing action-specific global information and reducing the disruption caused by irrelevant views.
Dual-Recommendation Disentanglement Network (DRDN) \cite{liu2023dual} applied a Specific Information Recommendation (SIR) to investigate a more suitable representation, striking a balance between action-specific and view-specific cues.
However, they still rely on additional information gained from view-specific features to obtain more complete action representation.
% Consequently, it is crucial to effectively aggregate the global features while concurrently focusing on the sparse yet vital detailed information in each view.
\begin{figure}[tb] 
  \centering
  \includegraphics[width=4in]{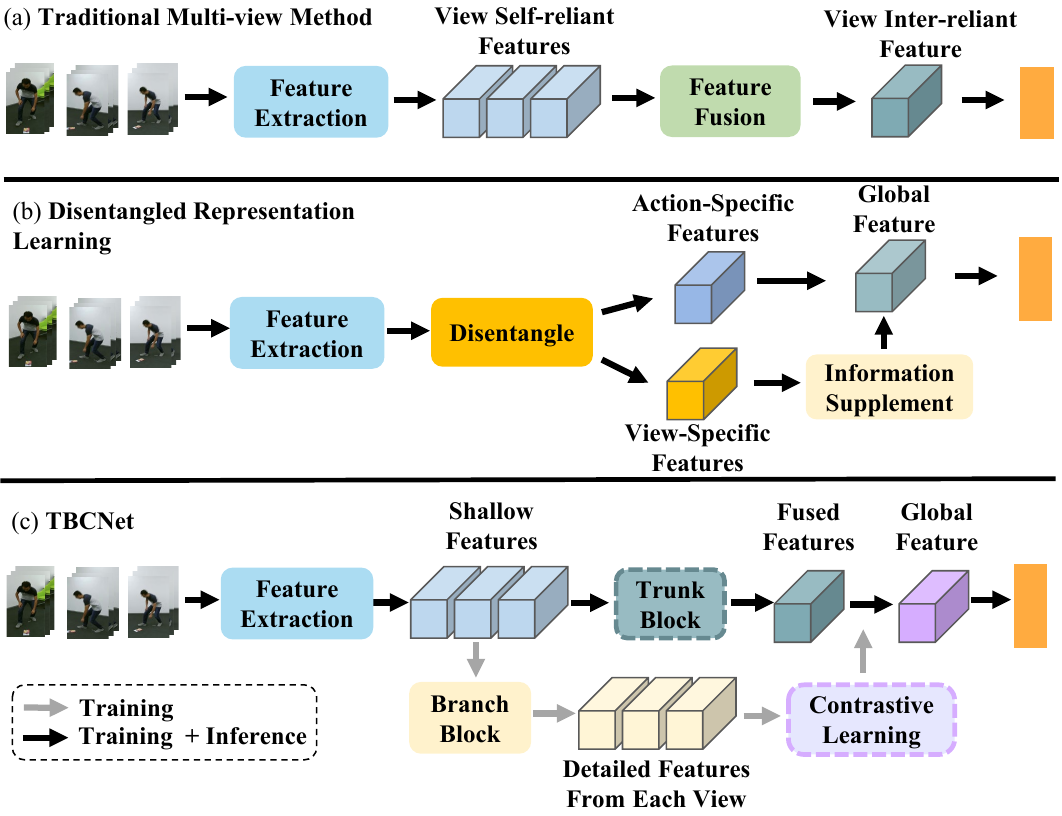}
  \caption{The architecture of different methods, including traditional multi-view methods, the method using disentangled representation learning and the proposed TBCNet.
  %After aggregating features from a global perspective in the trunk block, the proposed network subsequently leverages contrastive learning to augment the comprehensive representation by incorporating the details from each view in the branch block.
  }
  \label{img1}
\end{figure}

In the observation of objects within multi-view scenarios, human typically starts with the acquisition of global impression from a macro perspective, followed by a detailed exploration within each view to complement the comprehensive understanding.
Inspired by the hierarchical cognitive process, we propose a novel trunk-branch contrastive network (TBCNet) for RGB-based multi-view action recognition. The different architectural designs are depicted in Fig. \ref{img1}, unlike previous studies that carried out interactive fusion or action-view disentangling, the network first aggregates the multi-view features from a global perspective within the trunk block, and then applies the trunk-branch contrastive learning to promote fused features to absorb and complement pivotal details from each view within the branch block. 
This framework can obtain a more informative and comprehensive representation, and allows for the removal of the branch block during inference, reducing the actual running complexity.

\begin{figure}[tb]
\centering
  \subfigure[Multi-view deformable aggregation]
		{
			\begin{minipage}[b]{.45\linewidth}
				\centering
                \includegraphics[width=2.4in]{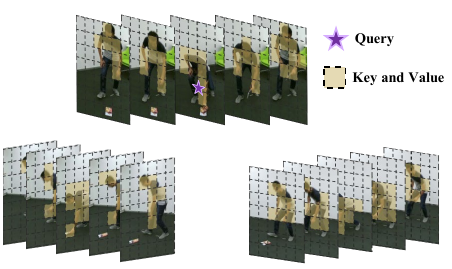}
            \end{minipage}
		}
  \subfigure[Trunk-branch contrastive learning]
		{
			\begin{minipage}[b]{.45\linewidth}
				\centering
    \includegraphics[width=2.4in]{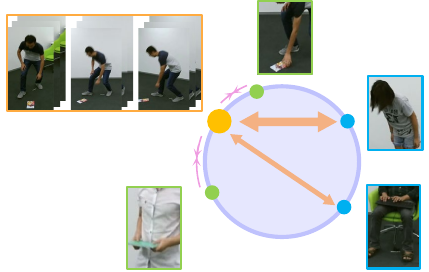}
  \end{minipage}
		}
  \caption{The schematic of the key components in the work. (a) Multi-view deformable aggregation. For a given query, the module attends to important deformed points in the entire multi-view sequence. (b) Trunk-branch contrastive learning. The fused feature (the yellow circle) employs contrastive learning to extract discriminative details from positive (green circles) and negative samples (blue circles).}
  \label{img2}
\end{figure}

In details, we propose a multi-view deformable aggregation in the trunk block to facilitate cross-view feature fusion, as shown in Fig. \ref{img2}(a).
Considering the critical role of complementary spatial information in multi-view settings, a global aggregation module is proposed into this module to aggregate and highlight significant spatial information at each frame.
%To avoid the confusion of the relative position of attention mechanism in cross-view context, 
And a composite relative position bias is designed for distinguishing the intra- and cross-view position of attention mechanism in multi-view scenes.
On the other hand, to generate a more holistic global representation, a novel trunk-branch contrastive loss is constructed between aggregated feature and refined behavioral information from the each view, as shown in Fig. \ref{img2}(b). 
Furthermore, by assigning two distinct weights for positive and negative samples, a weighted trunk-branch contrastive loss is proposed to emphasize subtle differences across classes and extract more valuable information. 
The main innovations of this article are delineated below:
\begin{enumerate}
\item We propose a trunk-branch contrastive network (TBCNet) for RGB-based multi-view action recognition, which leverages the trunk-branch contrastive learning to ensure that aggregated features absorb the significant details from each view, thereby capturing comprehensive representations.
And a weighted trunk-branch contrastive loss is proposed to augment the perception ability for subtle difference in diverse samples.

\item We propose a multi-view deformable aggregation (MVDA) to effectively fuse multi-view features with cross-view spatio-temporal correlations, 
where a global aggregation module (GAM) is proposed to emphasize the vital spatial information and a composite relative position bias (CRPB) is designed to capture intricate cross-view position information.
realize
\item We evaluate the proposed approach on four widely used datasets, including NTU-RGB+D 60, NTU-RGB+D 120, PKU-MMD and N-UCLA datasets. Results show that our method markedly outperforms other RGB-based approaches in both cross-subject and cross-setting protocols.
\end{enumerate}

%------------------------------------------------------------------------- 
\section{Related Works}

\subsection{Multi-View Action Recognition}

Given the prevalence of multi-view environments, researchers have devoted their efforts to action recognition in such scenarios.
Due to the superior spatial structure and robustness against illumination and background interference inherent in skeleton data, skeleton-based works \cite{Pan2023ViewNormalized,chi2022infogcn} had achieved excellent performance in multi-view action recognition.
To reduce the intra-class variances in skeleton data, a variance reduction framework \cite{Pan2023ViewNormalized} is proposed to generate a view-normalized skeleton and adjust the human pose according to the kinematic structure.
%By integrating spatio-temporal information of different modes, researchers achieve more accurate prediction results. 
Several studies incorporated multi-modality features into multi-view action recognition to enhance robustness, Cross-modality Aggregated Transfer (CAT) \cite{xu2021cross} was proposed to combine multi-view features and strengthens the global view-invariance features. 

%Liu et al. \cite{liu2023dual} proposed a Dual-Recommendation Disentanglement Network (DRDN) to resist the view-fuzzy noise via an adaptive cooperative representation between view and action components.
Many RGB-based methods were dedicated to extract view-invariant features.
Vyas et al. \cite{vyas2020multi} utilized cross-view prediction as an auxiliary task to learn view-invariant representations. 
Similarly, ViewCLR \cite{das2023viewclr} introduced a view-generator to produce latent viewpoint representations to generalize unseen camera viewpoints. 
Recent works \cite{liu2023dual,siddiqui2024dvanet} tried to introduce disentanglement learning in action recognition.
DVANet \cite{siddiqui2024dvanet} integrated learnable transformer decoder queries with supervised contrastive losses and was proven to yield uni-modal SOTA performance.
Since skeleton information requires additional acquisition processing, it may not be practical in multi-view scenes. Thus, we concentrate on RGB-based multi-view action recognition.

\subsection{Deformable Attention}
Deformable convolution \cite{zhu2019deformable} can dynamically attend to flexible spatial locations based on input.
To decrease the number of tokens in the vision transformer, a deformable attention transformer (DAT) \cite{Xia2022VisionTW} integrated this mechanism into the self-attention module, dynamically determining optimal positions of key-value pairs to extract more informative features. 
Following this, numerous studies had expanded deformable attention into video analysis, where the 3D deformable attention \cite{Kim2022CrossModalLW} was employed to adaptively model the spatio-temporal correlation. 
And some studies employed deformable attention on multi-view tasks,
MVDeTr \cite{hou2021multiview} adopted an introduced shadow transformer to attend varying positions to alleviate shadow-like distortions.
% Subsequent researches \cite{Liu2024MultiViewAC} employed the deformable attention to dynamically model the spatial feature relationship, showing consistent detection performance improvement.

Our study is the first to incorporate deformable attention into multi-view action recognition. To adapt fully to this task, we devise a global aggregation model that facilitates spatial fusion and propose a composite relative position bias to encode the multi-view positional condition.

\subsection{Contrastive Learning}
Contrastive learning had recently emerged as a prominent paradigm in representation learning, demonstrating promising performance advancements in computer vision \cite{Zhu2022BalancedCL}.
It emphasizes maximizing the similarities of positive pairs while minimizing that of negative pairs in a feature space during training.
Khosla et al. \cite{khosla2020supervised} extended the contrastive loss to effectively leverage label information in supervised learning, facilitating samples from the same class to converge into the cohesive embedding space.
%To get more meaningful negative samples, Kalantidis et al. \cite{kalantidis2020hard} proposed hard negative mixing strategies with the real-time computation.
Currently, contrastive learning is widely adopted in video understanding.
% The principal challenge pertains to the construction of suitable positive and negative sample pairs, ensuring that the feature representations learned are more comprehensible for subsequent tasks.
The Contrastive Multiview Coding (CMC) framework \cite{tian2020contrastive} sought to optimize the mutual information between diverse perspectives of an identical scene.
% To learn more discriminative feature representations, Wu et al. \cite{Wu2023Neighbor} proposed a novel neighbor-guided category-level contrastive learning to minimize the distance between the sample and its neighbors while maximizing the distance between the sample and other negative samples at the feature level. 
To enhance robustness against viewpoint alterations, Shah et al. \cite{shah2023multi}  employed hard negative samples to cultivate a more discerning information.%introduced a novel supervised contrastive learning framework and

In contrast to previous studies \cite{shah2023multi,siddiqui2024dvanet} that construct positive and negative sample pairs across views, our approach employs fused features as anchors while treats the details as positive and negative samples, implicitly promoting global features to absorb critical information from different views.

%------------------------------------------------------------------------- 

\section{Proposed Approach}

In this section, the overview of the proposed TBCNet is first introduced in Sec. \ref{sec3.1}.
Subsequently, we elucidate details about multi-view deformable aggregation and trunk-branch contrastive learning in Sec. \ref{sec3.2} and Sec. \ref{sec3.3}, respectively.
Finally, training and inference settings are given in Sec. \ref{sec3.4}.

\subsection{Overview of the Proposed Method}\label{sec3.1}

\begin{figure*}[tb]
\centering
  \includegraphics[width=5in]{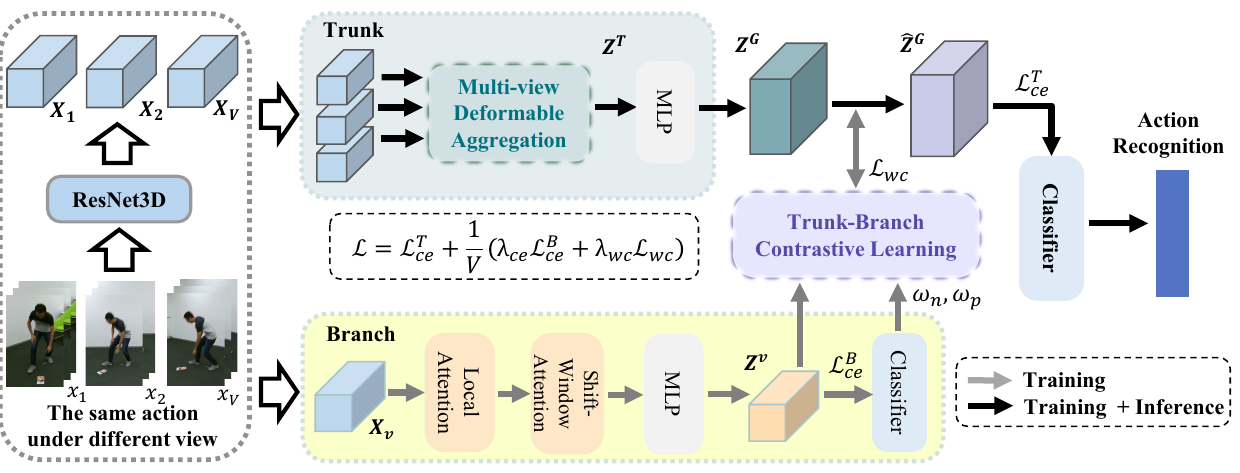}
  \caption{Outline of the proposed trunk-branch contrastive network. The trunk block employs multi-view deformable aggregation (MVDA) for the global feature fusion. And the trunk-branch contrastive learning facilitates the fused feature $\mathbf{Z}^G$ in absorbing the valid detailed information $\mathbf{Z}^B_v (v \in [1,V])$ from the branch block, yielding a comprehensive representation $\hat{\mathbf{Z}}^G$. The classifier in branch block assigns effective weights $w_n,w_p$ to contrastive samples, emphasizing subtle differences between samples.
  }
  \label{img3}
\end{figure*}

Contrary to previous works which conduct interaction and fusion after acquiring refined features, we proposed a novel trunk-branch contrastive network, its overall architecture is shown in Fig. \ref{img3}. %which firstly carries out multi-view fusion in the trunk block and then extracts the potentially detialed information from the branch block through contrastive learning, 
Initially, multi-view videos $(x_1, x_2,\ldots, x_V)$ are sequentially input into ResNet3D \cite{hara2018can} in a non-specific order to extract the shallow features $(\mathbf{X}_1, \mathbf{X}_2, \ldots, \mathbf{X}_V)$, where $\mathbf{X}_{v} \in \mathbb{R}^{T\times C\times H\times W}$ ($v \in [1, V]$). The parameter $V$ is the number of views, and $T$, $C$, $H$, and $W$ represent the number of frames, channels, height, and width, respectively.

After that, shallow features are fed into two distinct crucial modules: the trunk block and the branch block.
In the trunk block, shallow features are uniformly incorporated into the multi-view deformable aggregation to facilitate the feature amalgamation across views, and the output $\mathbf{Z}^T$ is then mapped to $\mathbf{Z}^G$ with a suitable dimension for the subsequent contrastive learning.
Meanwhile, $\mathbf{X}_{v}$ is input individually into the branch block, where vital details $\mathbf{Z}^B_{v}$ are extracted by local attention and shift-window attention \cite{Liu2021SwinTH}.
Then, $\mathbf{Z}^G$ integrates discriminative details $\mathbf{Z}^{B}_{v}$ via trunk-branch contrastive learning, with sample weights derived from the classifier in the branch block. Finally, the global representation $\mathbf{\hat Z}^{G}$ is input into a classifier for prediction.

\subsection{Multi-view Deformable Aggregation}\label{sec3.2}
\begin{figure*}[tb]
\centering
  \includegraphics[width=5.5in]{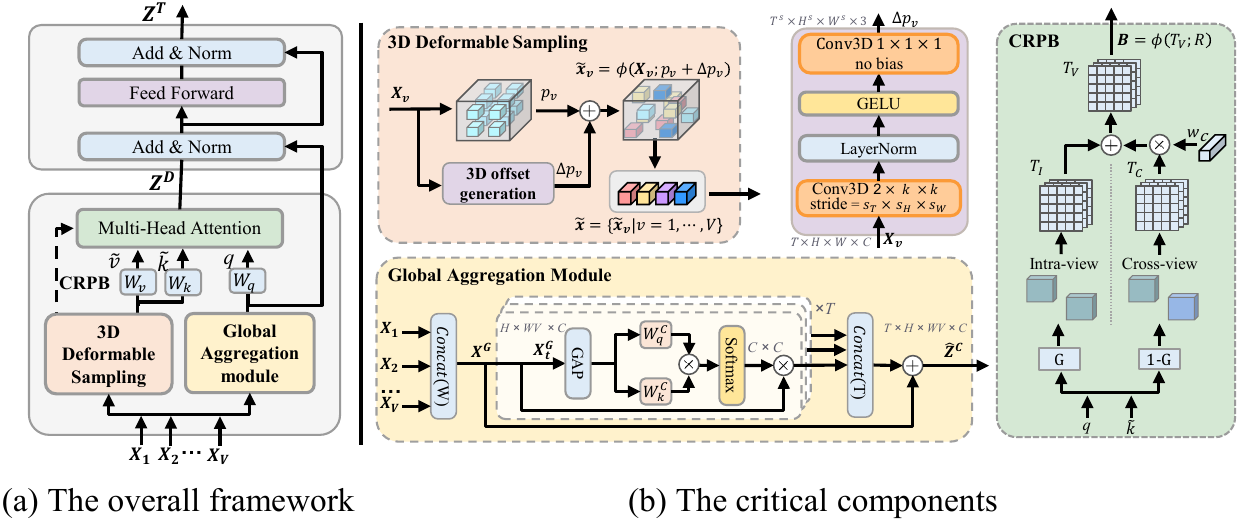}
  \caption{The illustration of the proposed multi-view deformable aggregation. %3D deformable sampling is used for generating the sampling feature set $\tilde{x}$ encompassing all views. The global aggregation module (GAM) employs the features with spatial concatenation as input, facilitating multi-view spatial compression and aggregation across channels. And composite relative position bias (CRPB) is designed to supplement intra- and cross-view position information.
  }
  \label{img4}
\end{figure*}

Inspired by DAT \cite{Xia2022VisionTW}, we construct the multi-view deformable aggregation (MVDA) for the multi-view feature fusion with adaptive spatio-temporal correlations. 
As shown in Fig. \ref{img4}, MVDA enhances the original deformable sampling of DAT by introducing the temporal information and incorporates two crucial components: the global aggregation module (GAM) and the composite relative position bias (CRPB). The former is designed to amplify vital spatial information in a global perspective, while the latter is utilized to capture the intricate position in the multi-view context.  

The overall framework of MVDA is shown in Fig. \ref{img4}(a). The keys $\tilde{k}$ and values $\tilde{v}$ are generated through valid sampling features $\tilde{x}$ from a 3D deformable sampling module. And queries $q$ are obtained from aggregated feature $\hat{Z}^C$ derived from the GAM. The attention module is formulated as:
\begin{equation}
\begin{aligned}
q&=\mathbf{\hat Z}^CW_{q}, \tilde{k}=\tilde{x}W_{k}, \tilde{v}=\tilde{x}W_{v},\\
\mathbf{Z}^D&={\mathrm{Softmax}}(q\tilde{k}^{T}/\sqrt{d}+ B(q,\tilde{k}))\tilde{v},
\end{aligned}
\end{equation}
where $W_{k}, W_{v}$ and $W_{q}$ are learnable projection matrix, $d$ is the number of channels, and $B (q,\tilde{k})$ represents the CRPB.
Finally, the output $\mathbf{Z}^T$ is derived from $\mathbf{Z}^D$ via the feed-forward network (FFN) with layer normalization (LN) and residual connection, represented as:
\begin{equation}
\mathbf{Z}^T=\mathrm{LN}\{(\mathrm{LN}(\mathbf{\hat{Z}}^C+\mathbf{Z}^D)) + \mathrm{FFN}(\mathrm{LN}(\mathbf{\hat{Z}}^C+\mathbf{Z}^D))\}.
\end{equation}

\subsubsection{3D Deformable Sampling}
With the same architecture as DAT \cite{Xia2022VisionTW}, the 3D deformable sampling is designed to adaptively assign crucial key-value pairs in the spatio-temporal context. 
As illustrated in Fig. \ref{img4}(b), for each input $\mathbf{X}_v$ ($v \in [1, V]$), reference points $p$ are derived from a uniformly distributed grid in the 3D space, with the sample number of $T^sH^sW^s$. Subsequently, a 3D offset generation module is used to calculate the local offset $\Delta p$ for these reference points.
Following this, prominent features are extracted at sampling locations of the reference point $p$ combined with its offset $\Delta p$, expressed by:
\begin{equation}
\begin{aligned}
&\Delta p_v=\theta_{\mathrm{offset}}(\mathbf X_v), v \in [1,V],\\
\tilde{x}=\{ \tilde {x}_v \mid &\tilde {x}_v =\phi(\mathbf X_v;p_v+\Delta p_v), v \in [1,V]\},
\end{aligned}
\end{equation}
where the function $\phi(\mathbf X;L)$ signifies that the feature $\mathbf X$ is sampled utilizing trilinear interpolation based on adjacent discrete coordinates at locations $L$.

\subsubsection{Global Aggregation Module} 
Under the context of multi-view feature fusion, the significance of spatial information complementarity is crucial. Thus, the global aggregation module is proposed to augment the aggregation of spatial features across views.

As shown in Fig. \ref{img4}(b), the module inputs spatially concatenated features $\mathbf X^G \in \mathbb{R}^{C\times T\times H\times WV}$ into a self-attention, where the head number of the self-attention equals the number of the frame $T$. In each attention head, and the split input $\mathbf X^G_t (t \in [1, T])$ undergoes a transformation via spatial global average pooling, being compressed to $\mathbf X^C_t \in \mathbb{R}^{C\times 1}$.
Then, the values and keys assigned from $\mathbf X^C_t$ are employed to calculate attention weight maps across channels. These maps represent the similarity among different channels across all views, facilitating the implicit aggregation of the global feature distribution.
Subsequently, $\mathbf X^G_t$ is multiplied by attention weight maps to re-weight action features in each frame, as formulated by:
\begin{equation}
\begin{aligned}
&\mathbf X^G=\mathrm{Concat}_W(\mathbf X_1,\ldots,\mathbf X_V), 
\mathbf X^C_t=\mathrm{GAP}(\mathbf X^G_t), \\
&\mathbf Z^C_t=\mathrm{Softmax}(\mathbf X^C_t W_q^{C} (\mathbf X^C_t W_k^{C})^{T}) \mathbf X^G_t, t \in [1, T],
\end{aligned}
\end{equation}
where $\mathrm{Concat}_W$ is the concatenating along the width dimension, and GAP means the spatial global average pooling. $\mathbf X^G_t$ is $t$-th vector of $\mathbf X^G$ in the time dimension. $W_q^{C}$ and $W_k^{C}$ are the learnable projection matrix.

Finally, the output is obtained by concatenating $\mathbf Z^C_t$ along the time dimension with a residual connection. The process can be expressed as:
\begin{equation}
\mathbf {\hat{Z}}^C=\mathrm{Concat}_T(\mathbf Z^C_1,\ldots,\mathbf Z^C_T)+\mathbf X^G,
\end{equation}
where $\mathrm{Concat}_T$ is the concatenating along the time dimension.

\subsubsection{Composite Relative Position Bias}

To separately learn the intra- and cross-view position information, we propose a composite relative position bias. The bias is constructed by an intra-view relative position bias table $T_I$ and a cross-view relative position bias table $T_C$. Both tables have a size of  $(2T-1,2H-1,2W-1)$. And a gating mechanism is introduced to assist the network differently access the two tables depending on the view condition under which attention pairs are entered.
%Given that the input of multi-view data is not contingent on the order of views, cross-view relative position can be encoded utilizing a singular table $T_C$. However
Meanwhile, to prevent potential located overlays across views, we design a learnable cross-view weight $w_{c}$ with the size of $ \genfrac (){0pt}{1}{v}{2}$ to weight the position encoding from different views. The process can be formulated as:
\begin{equation}
T_V=G(q,\tilde k)T_I+(1-G(q,\tilde k))w_{c}(q,\tilde k)T_C
\end{equation}
which $G(q,\tilde k)$ = 1 if query $q$ and key $\tilde k$ are in the same view else $G(q,\tilde k)$ = 0. $w_{c}(q,\tilde k)$ represents the cross-view weight in view combinations where $q$ and $\tilde k$ are located. 
Similar to DAT, 
%the relative position bias is calculated by a trilinear interpolation function $\phi(\cdot;\cdot)$ to ensure any continuous attention pairs exits a position information, 
the final composite relative position bias is calculated by a trilinear interpolation function $\phi(\cdot;\cdot)$, expressed as:$ B(q,\tilde k)= \phi (T_V;R_{q,\tilde k}) $,
where $R_{q,\tilde k}$ represent the relative displacements from $q$ to $\tilde k$.

\subsection{Trunk-branch Contrastive Learning}\label{sec3.3}
Generally, the subtle features inherent in given actions are profoundly affected by the perspectives from which they are observed. For example, a high viewpoint primarily emphasizes the direction and magnitude of trunk movement, whereas a low viewpoint focuses on leg alterations and their interactions with the ground. 
Inspired by supervised contrastive learning \cite{khosla2020supervised}, the trunk-branch contrastive loss is proposed to implicitly promote global features to absorb critical details from different views.

Meanwhile, it is crucial to pay attention to distinctions among similar categories. For instance, identifying the difference between `nodding' and `shaking the head' is more important than that between `nodding' and `wiping the face'. 
To this end, we incorporate weights for both positive and negative samples to construct a weighted trunk-branch contrastive loss, improving the network's ability to distinguish  between  actions  with subtle differences.

\begin{figure}[tb]
\centering
  \subfigure[Trunk-branch contrastive learning]
		{
			\begin{minipage}[b]{.4\linewidth}
				\centering
                \includegraphics[height=2.5cm]{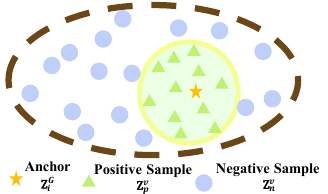}
            \end{minipage}
		}
  \hfill
  \subfigure[Weighted trunk-branch contrastive learning]
		{
			\begin{minipage}[b]{.55\linewidth}
				\centering
    \includegraphics[height=2.5cm]{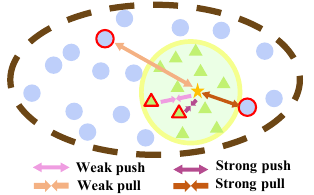}
  \end{minipage}
		}
  \subfigure[The relationship between the sample and the anchor]
		{
			\begin{minipage}[b]{.6\linewidth}
				\centering
    \includegraphics[height=2.5cm]{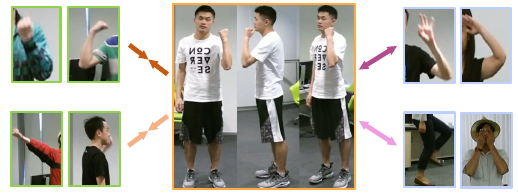}
  \end{minipage}
		}
  \caption{Illustration of the trunk-branch contrastive learning. The green area bounded by the yellow circle represents the feature space of the anchor's class.}
  \label{img5}
\end{figure}

\subsubsection{Trunk-branch Contrastive Loss}
Contrary to previous designs \cite{shah2023multi} for sample pairs, which utilize the current sample under a specific view as an anchor, and divide other samples from different views into positive and negative samples, we adopt global features $\mathbf{Z}^{G}$ that aggregates multi-view information as the anchor while choose the positive samples and negative samples from detailed features $\mathbf{Z}^{B}$.
%The contrastive learning is exclusively conducted between the global and perspective views, excluding any interactions between different views.
This design encourages the global feature to learn discriminative information and minimizes the background interference from the perspective. %that reflects behavioral variances 

For detailed feature $\mathbf{Z}_{v}^{B}$ under the view $v$, instances with the same class as anchor are designated as positive samples $\mathbf{Z}_{p}^{v}$, while those from other classes are classified as negative samples $\mathbf{Z}_{n}^{v}$, as shown in Fig. \ref{img5}. 
Based on the InfoNCE loss \cite{khosla2020supervised}, the proposed trunk-brank contrastive loss is defined as:
\begin{equation}
  \mathcal{L}_{c} =\sum_{v=1}^{V}
  \sum_{i=1}^{I}\frac{1}{|N_P|}\sum_{p=1}^{N_P}\mathcal{L}(v,i,p),
  \label{e_c1}
\end{equation}

\begin{equation}
    \begin{split}
    \mathcal{L}(v,i,p) =-\log
    \Bigg[\frac{\exp\left(
    \mathbf{Z}_{i}^{G} \cdot \mathbf{Z}_p^v
    /\tau\right)}
    {\exp\left(
    \mathbf{Z}_{i}^{G} \cdot \mathbf{Z}_p^v
    /\tau\right)
    +\sum_{n=1}^{N_N} \exp\left(\mathbf{Z}_{i}^{G} \cdot \mathbf{Z}_n^v/ \tau\right)} \Bigg], 
    \end{split}
    \label{e_c2}
\end{equation}
where $\mathbf{Z}_{i}^{G}$ is the fused feature of current anchor $i$ in the trunk block. $N_P$ and $N_N$ are the number of the positive and negative samples, respectively. 
The cosine distance $\mathbf{Z}^G \cdot \mathbf{Z}^v$ is applied to calculate the similarity between $\mathbf{Z}^G$ and $\mathbf{Z}^v$. $\tau$ is the temperature hyper-parameter.

\subsubsection{Weighted Trunk-branch Contrastive Loss}
Trunk-branch contrastive loss (Eq.\ref{e_c1} and Eq.\ref{e_c2}) is restricted to the sample level, where all positive or negative samples are equally considered, as shown in Fig. \ref{img5}(a).
To enable global features to discern more nuanced distinctions among diverse actions, the prediction probability from the classifier in the branch block is employed to establish two distinct weights for both positive and negative samples, as illustrated in Fig. \ref{img5}(b) and Fig. \ref{img5}(c).

In detail, the prediction probability $P(y_i|x_p)$ serves as a confidence estimate for the positive sample $x_p$, and it is chosen as the positive sample weight $w_{i,p}=P(y_i|x_p)$.
%Each positive sample is assigned a weight through the weighted summation for contrastive loss in each view. 
This manner can promote the network to extract more valuable information and reduce the impact of invalid information. %from samples that exhibit optimal recognition 
Concurrently, the negative sample weight is calculated by the exponential mapping of $P(y_i|x_n)$, reflecting the likelihood the negative sample $x_n$ is incorrectly classified as the anchor's label $y_i$, formulated as $w_{i,n}=\exp\left(P(y_i|x_n)\right)$. %, with ranging within [1,e].
By assigning higher weights to the hard negative samples, network can develop a more robust capability for fine-grained discrimination and focus on critical samples across diverse actions and views. The updated loss is given below:
\begin{equation}
  \mathcal{L}_{wc} =\sum_{v=1}^{V}
  \sum_{i=1}^{I}\frac{1}{\sum_{p=1}^{N_P}w_{i,p}}
  \sum_{p=1}^{N_P}w_{i,p} \mathcal{L}_w(v,i,p),
  \label{e_c3}
\end{equation}
the loss with negative sample weight is as follows:
\begin{equation}
    \begin{split}	
    \mathcal{L}_w(v,i,p) =-\log
    \Bigg[ \frac{\exp\left(
    \mathbf{Z}_{i}^{G} \cdot \mathbf{Z}_p^v
    /\tau\right)}
    {\exp\left(
    \mathbf{Z}_{i}^{G} \cdot \mathbf{Z}_p^v
    /\tau\right)
    +\sum_{n=1}^{{N_N}} \exp\left[w_{i,n}\left(\mathbf{Z}_{i}^{G} \cdot \mathbf{Z}_n^v/ \tau\right)\right]} \Bigg].
    \label{e_c4}
    \end{split}
\end{equation}

\subsection{Training and Inference}\label{sec3.4}
The loss of our network is composed of three parts. The cross-entropy loss $\mathcal\mathcal{L}_{ce}^{T}$ and $\mathcal{L}_{ce}^{B}$ are used for training trunk the branch blocks, respectively.
And contrastive loss $\mathcal{L}_{wc}$ is utilized to facilitate global features in effectively absorbing the subtle information of each view.
The overall loss is defined as:
\begin{equation}
    \mathcal{L}=\mathcal{L}_{ce}^{T} + \frac{1}{V}(\lambda_{ce}\mathcal{L}_{ce}^{B} + \lambda_{wc}\mathcal{L}_{wc}),
    \label{e_t1}
\end{equation}
where $\lambda$ means the weights for different loss $\mathcal{L}$.

Noteworthy, the trunk block exhibits limited discrimination capabilities during early training, potentially causing distortions when directly introducing contrastive learning. Thus, we adopt a two-stage training strategy. 
In the initial phase, only $\mathcal\mathcal{L}_{ce}^{T}$ and $\mathcal{L}_{ce}^{B}$ are used to train the trunk and branch blocks independently.
Subsequently, contrastive loss is incorporated to facilitate information transmission. During inference, it is feasible to remove the branch block and solely utilize the trunk block.

%------------------------------------------------------------------------- 
\section{Experimental results and analysis}
In the section, we initially present four popular datasets in Sec. \ref{sec4.1}, followed by implementation details in Sec. \ref{sec4.2}. In Sec. \ref{sec4.3}, we perform ablation studies to verify the effectiveness of each component. Sec. \ref{sec4.4} compares the proposed method with other state-of-the-art (SOTA) methods. Sec. \ref{sec4.5} offers qualitative analysis and visualization to further assess the performance of the method. Lastly, Sec. \ref{sec4.6} outlines current limitations.

\subsection{Datasets}\label{sec4.1}
The proposed method is evaluated on four widely used multi-view action recognition datasets: NTU-RGB+D 60 \cite{shahroudy2016ntu}, NTU-RGB+D 120 \cite{liu2019ntu}, PKU-MMD \cite{liu2017pku}, Northwestern-UCLA Multi-view Action (N-UCLA) \cite{Wang2014CrossViewAM}. They consist of multi-view video groups captured by three independent cameras. 

\textbf{NTU-RGB+D 60} provides 56,880 samples collected from 40 subjects and captured from three viewpoints. The dataset contains 60 different action classes including daily, mutual, and health-related actions. For evaluation, we adhere to the cross-subject (CS) and cross-view (CV) protocols. For CS protocol, the data from 20 subjects are used for training, with the rest reserved for testing. In CV evaluation, samples from Camera 1 are utilized for testing, while those from Camera 2 and Camera 3 are employed for training.

\textbf{NTU-RGB+D 120} extends NTU-RGB+D 60 by adding 60 classes, consisting of 114,480 samples from 106 different subjects with 155 distinct viewpoints. We follow cross-subject (CS) and cross-setup (CSet) protocols for evaluation. In CS and CSet evaluations, data are split into training and testing groups based on the subject ID and setup ID, respectively.

\textbf{PKU-MMD} contains 1076 long, untrimmed videos from 3 cameras and involves 51 actions performed by 66 subjects. It also follows the cross-subject (CS) and cross-view (CV) protocols. In CS evaluation, the dataset is divided into training and testing groups which consist of 57 and 9 subjects respectively. For CV evaluation, videos from Camera 2 and Camera 3 are chosen for training set and the rest are for testing set.

\textbf{N-UCLA} collects about 1,500 videos and involves 10 actions performed by 10 actors. Similarly, the evaluation protocol applies cross-subject (CS) and cross-view (CV). In CS evaluation, videos from 9 subjects are used for training, and the rest are used for testing. For CV evaluation, videos from the first two cameras are set for training and the others for testing.

\subsection{Implementation Details}\label{sec4.2}
The proposed model is implemented using the PyTorch framework on Linux with two 4090 24GB GPUs. AdamW optimizer is adopted with a learning rate of $10^{-5}$ and weight decay of $10^{-5}$. The batch size is 16 for NTU-RGB+D 60, NTU-RGB+D 120 datasets and PKU-MMD while 8 for N-UCLA dataset. Each batch contains center-cropped videos with 16 fixed frames at 224×224 resolution. The pre-trained ResNet3D \cite{Feichtenhofer2018SlowFastNF} serves as the backbone network. 
Our method exclusively utilizes RGB data during both training and testing.
During the initial 150 training epochs, the model does not incorporate contrastive learning, with $\lambda_{ce}$ and $\lambda_{wc}$ set to 1 and 0, respectively. In the subsequent 100 epochs, these values are adjusted to 0.5 and 0.25.
Given that the proposed network requires at least two views as inputs, it concurrently loads two identical sets of test data in the CV benchmark.

\begin{table}[t]
\centering
{\fontsize{10}{10}\selectfont}
\small 
\caption{Ablation study of MVDA on NTU-RGB+D 60. GAM is the global aggregation module while DA is the self-attention with 3D deformable sampling. RPB is the relative position bias while CRPB is the proposed composite relative position bias. 
}
\label{tab1}
\begin{tabular}[width=0.8\textwidth]{ccccc}
    \toprule
    ResNet3D& GAM & DA & Position Encoding & CS(\%) \\ \midrule
    $\checkmark$ &$\times$     &$\times$     & $\times$           & 88.9\\
    $\checkmark$ &$\checkmark$ &$\times$ &$\times$              & 91.5\\
    $\checkmark$ &$\checkmark$ &$\checkmark$ &$\times$             & 93.7\\\midrule  
    $\checkmark$ &$\checkmark$ &$\checkmark$ & RPB                & 93.8\\
    $\checkmark$ &$\checkmark$ &$\checkmark$ & CRPB             & 94.0\\
    \bottomrule
\end{tabular}
\end{table}

\subsection{Ablation Studies}\label{sec4.3}
To assess the significance of key components in TBCNet, the ablation analysis is conducted under the CS protocol on NTU-RGB+D 60. ResNet3D is used as the baseline in these experiments.

\subsubsection{Effectiveness of MVDA} 

Table \ref{tab1} shows the results of ablation experiments for MVDA.
Contrary to prior methods \cite{wang2018dividing, ULLAH2021321} that conducted pairwise fusion without dividing space and time, the GAM highlights vital spatial information from a global perspective, and boosts performance by 2.6\%.%, which proves the significance of integrating multi-view spatial information.
To facilitate valid spatio-temporal fusion across diverse view, we apply the 3D deformable attention in multi-view contexts, enhancing the accuracy by 2.2\%.  
Compared to the relative position encoding, CRPB considers the variance between intra- and inter-view positions and employs an additional weight to learn the diverse information across different views, further enhancing performance by 0.3\%.%, indicating distinguishing multi-view positions is practical.

\begin{figure}[t]
\centering
  \includegraphics[width=4.2in]{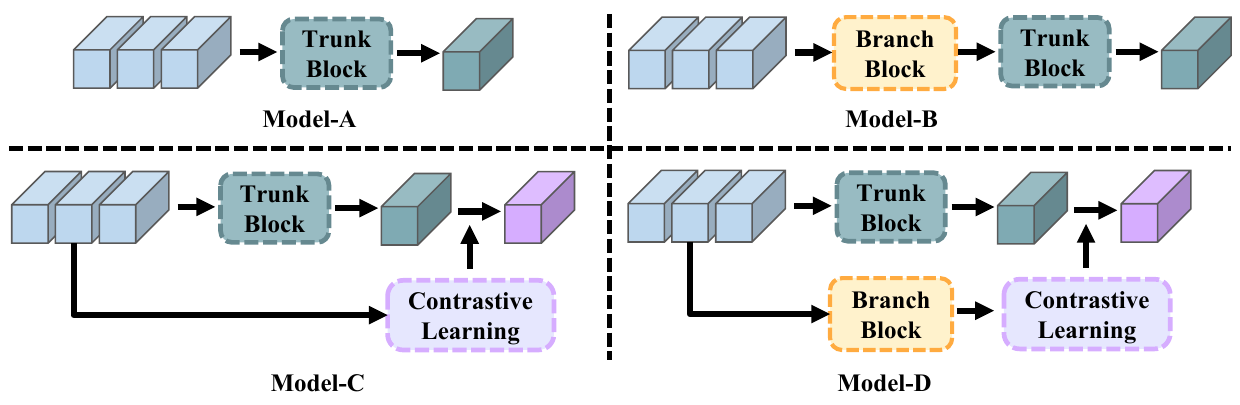}
  \caption{Four different models were used for evaluating the effectiveness of the trunk-branch framework in ablation experiments. Model-D is the proposed architecture.
  }
  \label{img6}
\end{figure}

\subsubsection{Effectiveness of Trunk-branch Contrastive Framework}
To evaluate the effectiveness of the designed framework, we introduce four different architectures for ablation experiments. Fig. \ref{img6} shows these models: Model-A only utilizes the trunk block while Model-B applies a series connection between trunk and branch blocks. Model-C inputs shallow features directly into contrast learning, and Model-D is the proposed framework.
%the blue box is the multi-view shallow feature extracted by ResNet3D, the green box is the fused feature while the purple is the transformed fused feature via contrastive learning. 

To illustrate the impact of simply integrating contrast learning, we firstly implement the comparison between Model-A and Model-C. As shown in Table \ref{tab2}, Model-C has a 0.2\% accuracy drop relative to Model-A. This decline occurs because the contrastive learning in Model-C fails to provide additional useful information and instead disrupts the information reinforcement from MVDA. Model-D introduces the branch block with fine-grained feature extraction into contrastive learning, enabling contrastive learning to positively contribute to the fused feature, leading to the performance improvement.
\begin{table}[t]
\centering
{\fontsize{10}{10}\selectfont}
\small
\caption{Ablation study of the different framework on NTU-RGB+D 60. The class of model is shown in Fig. \ref{img6}}
\label{tab2}
    \begin{tabular}{cccc}
    \toprule
    Model & Param(M) & FLOPs(G) &CS(\%)\\ \midrule 
    Model-A & 4.52 & 38.82 &94.0\\
    Model-B & 6.63 & 39.44 &95.3 \\
    Model-C & 4.52 & 38.82 &93.8\\
    Model-D & 6.77 & 40.06 &95.6\\
    \bottomrule
\end{tabular}
\end{table}

On the other hand, Model-B and Model-D introduce the branch block in distinct ways.
Specifically, contrastive learning in Model-D effectively accentuates the disparities between positive and negative samples, enabling the model to better integrate salient features from each view. It leads to a 0.3\% accuracy improvement over Model-B.
Remarkably, the branch block can be eliminated at inference duo to the information transmission relied on the loss. %of the branch block in Model-D  during training. Consequently, %
This strategy enables augmentation for the trunk block without adding complexity, especially using more complex models in the branch block. As a result, the proposed framework has better performance than Model B, with the same number of parameters as Model-A (as shown in Table \ref{tab2}) in practical application. It demonstrates the superiority of our design.

\subsubsection{Effectiveness of Trunk-branch Contrastive Learning}
The model is incrementally modified by adjusting individual components or losses to assess the effectiveness of the proposed contrastive learning.
\begin{table}[H]
\centering
{\fontsize{10}{10}\selectfont}
\small
\caption{Ablation study of the effectiveness of branch block and contrastive learning on NTU-RGB+D 60. LA is the local attention while SWA represents the shift-window attention. $w_n$ is the negative sample weight while $w_p$ is the positive sample weight. }
\label{tab3}
\begin{tabular}{ccccccc}
    \hline
    Trunk & \multicolumn{2}{l}{\begin{tabular}[c]{@{}c@{}}Branch Block\end{tabular}}  & \multicolumn{3}{l}{\begin{tabular}[c]{@{}c@{}c@{}}Contrastive Loss\end{tabular}}  & \multirow{2}{*}{CS(\%)} \\ \cline{2-6}
     Block & LA & SWA & $\mathcal{L}_{c}$ & $w_p$ & $w_n$ & \\\hline
    $\checkmark$ &$\times$     &$\times$&$\times$   & $\times$& $\times$ & 94.0\\
    $\checkmark$ &$\checkmark$ &$\times$&$\checkmark$    &$\times$     & $\times$  & 94.9\\
    $\checkmark$ &$\checkmark$ &$\checkmark$&$\checkmark$ & $\times$ & $\times$  &95.6\\\hline
    $\checkmark$ &$\checkmark$ &$\checkmark$&$\checkmark$ & $\checkmark$ & $\times$ & 95.8\\
    $\checkmark$ &$\checkmark$ &$\checkmark$&$\checkmark$ & $\checkmark$ & $\checkmark$ &96.3
    \\\hline
\end{tabular}
\end{table}

As shown in Table \ref{tab3}, integrating local attention (LA) and shift-window attention modules (SWA) into the branch block improves performance by 0.9\% and 0.7\%, respectively. Their incorporation enriches global features with more informative details through contrastive learning.
Meanwhile, the application of positive sample weights highlights valuable instances with an enhancement of 0.2\%.
Conversely, negative sample weights enable the network to better distinguish between similar samples, resulting in a 0.5\% improvement.

%We first explore the impact of intricate feature extractor for contrastive learning.
%Meanwhile, we design positive sample weights and negative sample weights to amplify the unique impact of distinct samples and highlight valuable instances, enhancing the network's discriminative capacity. indicating that prioritizing high-accuracy samples enhances learning quality.as shown the rows 4-5 of Table \ref{tab3}
% Totally, the incorporation of contrastive loss and branch block enables global features to emphasize and learn salient features in each view. This results in a 2.3\% improvement in accuracy, thereby validating the efficacy of our proposed method. 

\subsection{Comparisons with the State-of-the-Arts}\label{sec4.4}

\subsubsection{NTU-RGB+D 60 and NTU-RGB+D 120} 

We conduct a thorough comparison for the proposed method and SOTA approaches using various modalities on NTU-RGB+D 60 and NTU-RGB+D 120. The results are presented in Table \ref{tab4}. 

\begin{table}[H]
\centering
{\fontsize{10}{10}\selectfont}
\small
\caption{Comparison to SOTA methods on NTU-RGB+D 60 and NTU-RGB+D 120. Bold text indicates the best performance, underlining indicates the second best performance.}
\label{tab4}
\setlength{\tabcolsep}{2.5pt}
\begin{tabular}{ccccccc}
\hline
\multirow{2}{*}{Modality}& \multirow{2}{*}{Method}& \multirow{2}{*}{Venue} & \multicolumn{2}{c}{NTU 60} & \multicolumn{2}{c}{NTU 120} \\ \cline{4-7}
&    &    & CS(\%)    & CV(\%)   & CS(\%)   & CSet(\%) \\ \hline
\multirow{6}{*}{Skeleton}
& InfoGCN \cite{chi2022infogcn}   & CVPR '22& 93.0    & \underline{97.1}  & \underline{89.9}  & \underline{91.1} \\
& 3Mformer \cite{wang20233mformer} & CVPR '23& \textbf{94.8}  & \textbf{98.7} & \textbf{92.0}    & \textbf{93.8} \\ 
& VaRe \cite{Pan2023ViewNormalized} & TCSVT '23 & 92.0& 96.7& 87.6&89.4\\ 
& STHG-DAN \cite{Wu2024STHB}& PR '24&91.2 &96.5 &88.7&89.8\\ 
%& SPIANet \cite{Yin2024SpatiotemporalPI}& PR '24&92.8 &96.8 &89.2&90.4\\ 
& MV-TSHGNN \cite{Ma2024MultiViewTH} &TIP '24 &92.9 &96.8 &89.4&90.8\\ 
& LG-SGNet \cite{Wu2025LocalAG}& PR '25&\underline{93.1} &96.7 &89.4&91.0\\ 
\hline
\multirow{4}{*}{\begin{tabular}[c]{@{}c@{}}RGB+\\ Skeleton\end{tabular}} 
& PoseC3D \cite{duan2022revisiting}  & CVPR '22& \textbf{97.0}  & \textbf{99.6}  & \textbf{95.3}  & \textbf{96.4}\\
%& STAR \cite{ahn2023star} & WACV '23& 90.3  & 92.7 &90.3&92.7 \\
& MMNet \cite{MMNet2023}& TPAMI '23& 94.2 & 97.8  &90.3&92.1\\
& 3DA \cite{Kim2022CrossModalLW} & ICCV '23& 94.3  & 97.9  & 90.5  & 91.4\\
& $\pi$ -ViT \cite{reilly2024just}& CVPR '24& \underline{96.3}  & \underline{99.0} &\underline{95.1}  & \underline{96.1}\\
 \hline
\multirow{7}{*}{RGB}
& DANet \cite{wang2018dividing}   & ECCV '18& 88.1  & 84.2 &   -&   -\\
& CVAM \cite{vyas2020multi} & ECCV '20& 82.3  & 86.3 &   - &   -\\
& DRDN \cite{liu2023dual}  & TIP '23 & 91.2  & 92.9 &   - &   -\\ 
& ViewCLR \cite{das2023viewclr}  & WACV '23& 89.7  & 94.1  & 84.5  & 86.2\\
& ViewCon \cite{shah2023multi}   & WACV '23& 91.4  & \underline{98.0}  & 85.6  & 87.5\\
& DVANet \cite{siddiqui2024dvanet}  & AAAI '24& \underline{93.4}  & \textbf{98.2}&\underline{90.4}  & \underline{91.6} \\
\cline{2-7} 
& TBCNet (Ours)& -  & \textbf{96.3} & 97.1 & \textbf{91.5} & \textbf{92.9}\\ \hline
\end{tabular}
\end{table}

On the NTU-RGB+D 60 in the CS protocol, the method outperforms all unimodal SOTA methods, achieving a 2.9\% improvement over DVANet \cite{siddiqui2024dvanet}. 
DRDN \cite{liu2023dual} and DVANet \cite{siddiqui2024dvanet} enforce disentanglement for action and view features, exhibiting excellent performance. Unlike these methods, our method initially aggregates global features, then supplements detailed information in each view through contrastive learning. 
It enables TBCNet to learn discriminative and comprehensive features, achieving superior classification results.
MVDA is designed for global aggregation and cross-view association, making it less suited for single-view inputs during testing of the CV protocol. Despite this limitation, the model achieves 97.1\% accuracy, surpassing most existing algorithms and proving robustness to viewpoint shifts.

On the larger NTU-RGB+D 120, our approach demonstrates superior performance compared to DVANet which surpasses the other RGB-based method by over 4\% in accuracy. Specifically, our method achieves further improvements of 1.1\% and 1.3\% under CS and CSet standards respectively, showing superior generalization and adaptability to subject and background variations.
Under the CSet setting, testing data include videos from three distinct perspectives, MVDA can effectively amalgamate cross-view features and capture richer multi-view features.

%The MVDA in the trunk block enhance the convergence of spatial features by the global aggregation module and conduct the cross-view spatial-temporal aggregation via the deformable attention with composite relative position bias.

%Notably, the competitive advantage of our method over skeleton-based methods is less pronounced on the NTU-RGB+D 120 compared to the NTU-RGB+D 60. This could be attributed to the diverse environmental settings of the NTU-RGB+D 120, which encompasses a large number of outdoor scenes heavily affected by lighting conditions. RGB-based methods are more susceptible to such lighting variations than skeleton-based methods.
% In such environments, the RGB-based methods encounter the limitation due to the susceptibility of RGB data to interference from lighting variations compared to skeleton-based methods. 
% However, when compared to DVANet, which surpasses the other RGB-based method by over 4\% in accuracy, our method achieves an improvement of 1.1\% and 1.3\% under CS and CSet standards, respectively.

\begin{table}[H]
\centering
{\fontsize{10}{10}\selectfont}
\small
\caption{Comparison to SOTA methods on PKU-MMD dataset. Bold text indicates the best performance, underlining indicates the second best performance.}
\label{tab5}
\setlength{\tabcolsep}{8pt}
\begin{tabular}{ccccc}
\hline
Modality & Method       & Venue      & CS(\%)   & CV(\%)    \\ \hline
\multirow{4}{*}{\begin{tabular}[c]{@{}c@{}}RGB+\\Depth\end{tabular}} 
&TSM \cite{Lin2018TSMTS}& TPAMI'20& \underline{92.1}  & \underline{93.2}\\
&SC-CNN \cite{Ren2020Seg}& NeuCom'20& 91.7  & 92.6\\
& MMINet \cite{Cheng2022Spatial-Temporal}  & Access'22& \textbf{93.6}  & \textbf{94.2} \\
&J-CMCB \cite{Cheng2022}& TCSVT'22& 90.4  & 91.4\\
\hline
\multirow{2}{*}{RGB} 
&DVANet \cite{siddiqui2024dvanet}& AAAI ’24   & \underline{95.8} & \textbf{95.3} \\ \cline{2-5}
& TBCNet (Ours)  &    - & \textbf{96.4} & \underline{93.6} \\ \hline
\end{tabular}
\end{table}

\subsubsection{PKU-MMD}
Table \ref{tab5} shows the accuracy of our network compared with the existing SOTA methods on the PKU-MMD dataset.
DVANet \cite{siddiqui2024dvanet} employs contrastive learning among different views or actions to learn disentangled view and action representations. In contrast, we construct a trunk-branch contrastive loss between the global feature and refined features from various views to promote global feature to absorb vital details. This approach results in a 0.5\% improvement over DVANet, proving its efficacy.
But our method achieves relatively inferior results under the CV protocol due to the same testing configuration as the NTU-RGB+D 60 in the PKU-MMD dataset.

\begin{table}[H]
\centering
{\fontsize{10}{10}\selectfont}
\small
\caption{Comparison to SOTA RGB-based methods on N-UCLA dataset. Bold text indicates the best performance, underlining indicates the second best performance. }
\label{tab6}
\setlength{\tabcolsep}{10pt}
\begin{tabular}{cccc}
\hline
Method       & Venue      & CS(\%)   & CV(\%)    \\ \hline
DANet \cite{wang2018dividing}& ECCV '18   & 92.1 & 86.5 \\
CVAM \cite{vyas2020multi}& ECCV '20   & 87.5 & 83.1 \\
Conflux LSTM \cite{ULLAH2021321} & NeuCom '21 &   -  & 88.9 \\
DRDN \cite{liu2023dual}        & TIP ’23  & 92.1  & 93.9\\
ViewCLR \cite{das2023viewclr}      & WACV ’23   &   -  & 89.1 \\
ViewCon \cite{shah2023multi}      & WACV ’23   &   -  & 91.7 \\
DVANet \cite{siddiqui2024dvanet}       & AAAI ’24   & \underline{94.4} & \textbf{96.5} \\ \hline 
TBCNet (Ours)   &    -       & \textbf{95.5} & \underline{95.7} \\ \hline
\end{tabular}
\end{table}

\subsubsection{N-UCLA}
Table \ref{tab6} displays outcomes in our method and other RGB-based methods on N-UCLA.
In the designed weighted trunk-branch contrastive loss, %the network leverages samples with varying degrees of emphasis, prioritizing beneficial information. 
the positive sample weight emphasizes high-quality samples while the negative sample weight highlights slight differences among similar samples. 
Consequently, our approach achieves optimal performance of 95.5\% in the CS protocol on the N-UCLA with limited training data. Despite its limited effectiveness under the testing conditions in the CV protocol, it delivers a respectable accuracy by 95.7\%, surpassed only by DVANet.

\begin{figure*}[tb]
\centering
  \includegraphics[width=5.5in]{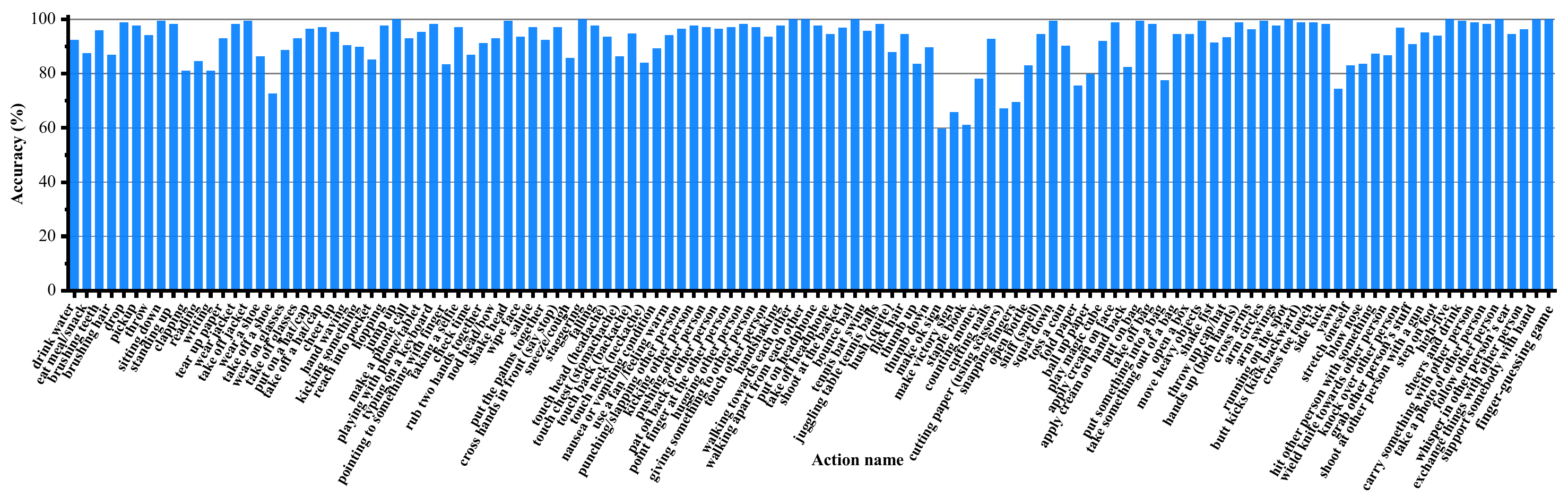}
  \caption{Classification accuracy for each action on NTU-RGB+D 120. Results are derived from the trunk block without using contrastive learning in cross-setup (CSet) protocol. %The average accuracy achieved is 91.97\%.
  }
  \label{img7}
\end{figure*}

\subsection{Qualitative Analysis and Visualization} \label{sec4.5}
% In this section, we first present the qualitative analysis of MVDA. Subsequently, trunk-branch contrastive learning is evaluated by comparing the differences observed before and after its introduction.

\subsubsection{Qualitative analysis for MVDA} 
Fig. \ref{img7} presents accuracy results of the trunk block on the NTU-RGB+D 120 using CSet protocol. 
Due to large observable areas provided by two-person interaction actions, MVDA can capture and aggregate valuable information, achieving over 95\% accuracy on these actions like `walking towards each other'.
However, for actions with minor movements like `staple book', partial views often fail to offer valid features, which hampers reliable consolidation in MVDA. Therefore, to obtain more overall information, it is essential to capture intricate details and mitigate the adverse impact of invalid views.

\begin{figure}[tb]
\centering
  \subfigure[Accuracy differences]
		{
			\begin{minipage}[b]{.26\linewidth} 
				\centering
               \includegraphics[height=4.6cm]{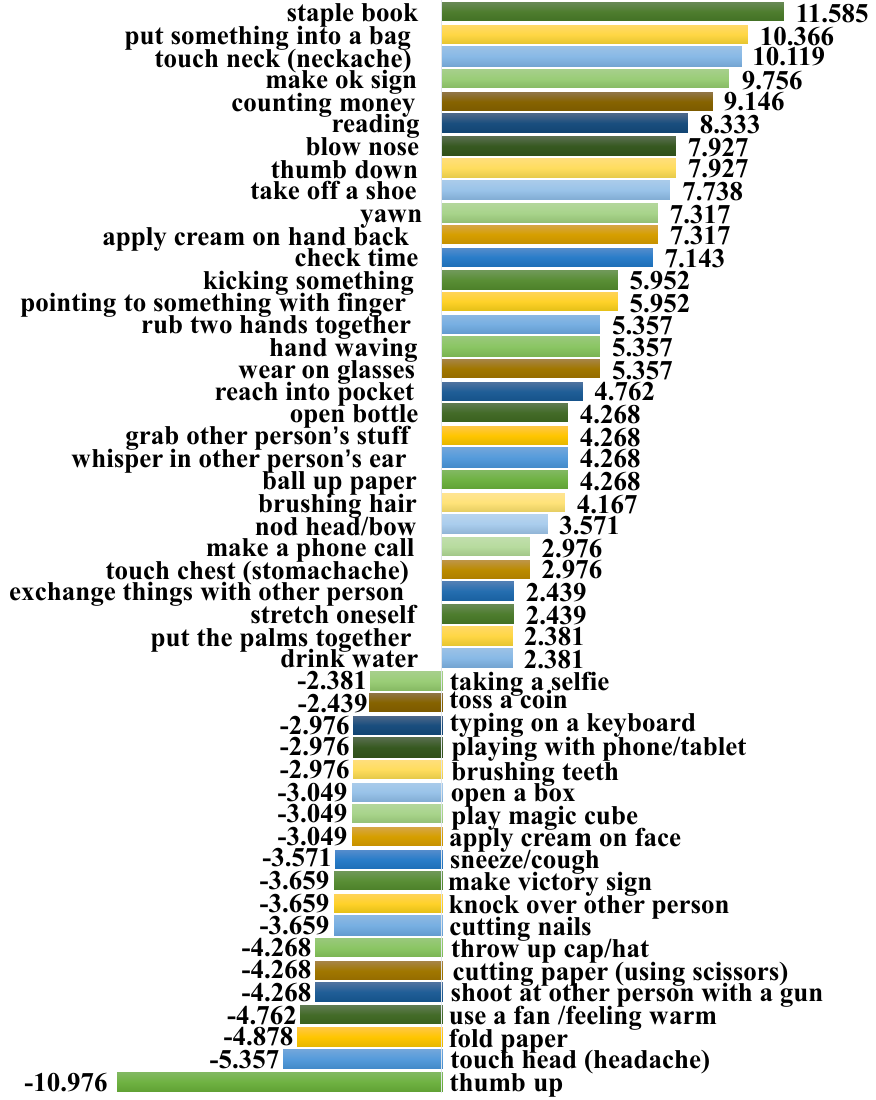}
            \end{minipage}
		}
  \hfill
  \subfigure[Trunk block]
		{
			\begin{minipage}[b]{.32\linewidth}  
				\centering
    \includegraphics[height=3.6cm]{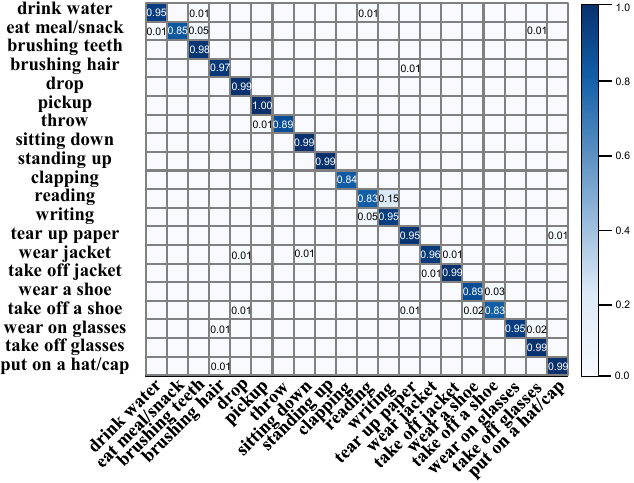}
  \end{minipage}
		}
  \hfill
  \subfigure[TBCNet]
		{
			\begin{minipage}[b]{.32\linewidth}  
				\centering
    \includegraphics[height=3.6cm]{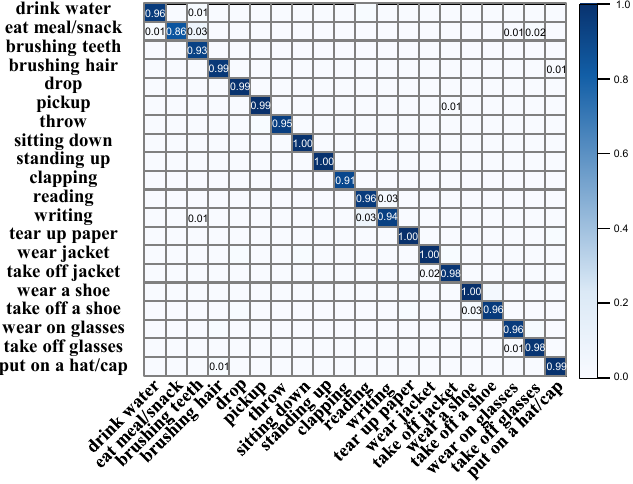}
  \end{minipage}
		}
  \caption{Illustration for the qualitative analysis in trunk-branch contrastive learning. (a) lists actions that recognition rate changes by more than 2\% before and after incorporating contrastive learning. 
  %The average accuracy of the complete network is 92.88\%. 
  (b) and (c) are confusion matrices for the first 20 classes on NTU-RGB+D 60 under the CS protocol, obtained from trunk block and TBCNet, respectively.}
  \label{img8}
\end{figure}

\subsubsection{Qualitative analysis for Trunk-branch contrastive learning}
Fig. \ref{img8}(a) depicts the alterations with the accuracy shift exceeding 2\% after introducing contrastive learning for NTU-RGB+D 120 in CSet protocol.
The integration of branch block and contrastive learning enables the network to focus on the nuanced representations of each view, which results in notable accuracy improvements for fine-grained actions like 'staple book' and 'yawn'.
Due to disparate upper limb movements in the dataset, overfitting occurs in consistent behavioral details captured by the branch block, leading to significant accuracy degradation in ‘thumb up'.
%Specifically, while some participants fully extended their arms, others maintained a bent position. Despite increased errors in a part action classes, there is an overall improvement in accuracy by 0.9\%.

Fig. \ref{img8} also presents confusion matrices of the first 20 classes on NTU-RGB+D 60. Fig. \ref{img8}(b) reveals the effective aggregation ability of the trunk block, which achieves over 95\% accuracy on most classes. 
However, significant discrimination errors are exhibited in specific actions like ‘wearing a jacket' and ‘taking off a shoe'. %certain actions such as ‘clapping' and ‘reading'  accuracy slightly below 85\%, with 
Fig. \ref{img8}(c) shows results of TBCNet, where all classes achieve over 95\% except for ‘eat meal/snack'.
This is because the branch block and contrastive learning offer valid details while the weighted trunk-branch contrastive loss promotes better discrimination among similar classes, effectively reducing misjudgments.

\begin{figure}[t]
\centering
  \includegraphics[width=5.4in]{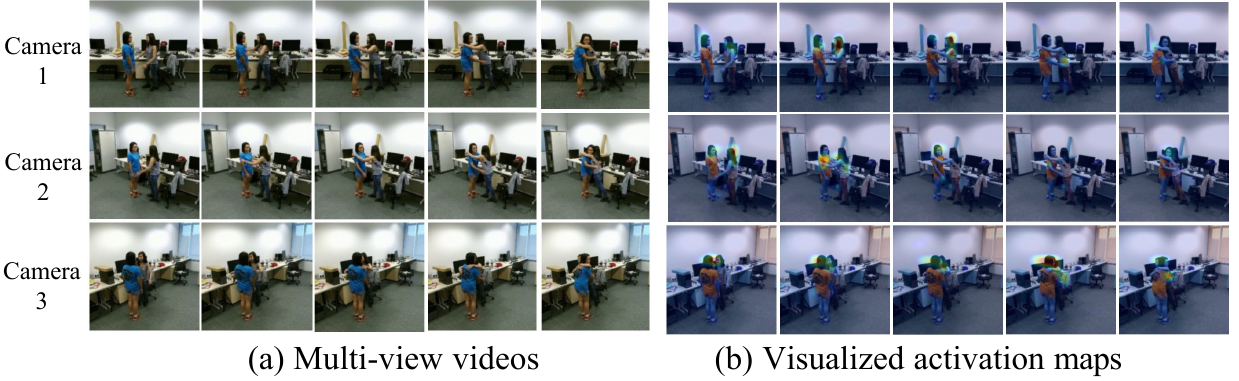}
  \caption{Visualized activation maps for `hugging other person'. 
  }
  \label{img9}
\end{figure}

\subsubsection{Visualization}
% To further evaluate our approach, we visualize the activation maps under multi-view video sequences, as shown in Fig. \ref{img9}.
%Since the network relies solely on RGB videos rather than skeletal data, it struggles to accurately capture full arm movements.However, 
To further evaluate our approach, we visualize activation maps under multi-view video sequences. %As shown in activation maps of Fig. \ref{img9},
As shown in Fig. \ref{img9}, the proposed method effectively focuses on the nuanced action information associated with the hand, upper arm, and head across various temporal positions and views. % in the entire multi-view sequence. 
And the network highlight the relative positioning of two individuals and hand-body contact in Camera 1 and Camera 2. In contrast, it emphasizes head interaction and the positioning of both upper limbs in Camera 3. It underscores the efficacy in capturing diverse visible information from various perspectives.

\begin{figure}[bt]
\centering
  \subfigure[ResNet3D]{
    \begin{minipage}[b]{.3\linewidth}
        \centering
        \includegraphics[height=3cm]{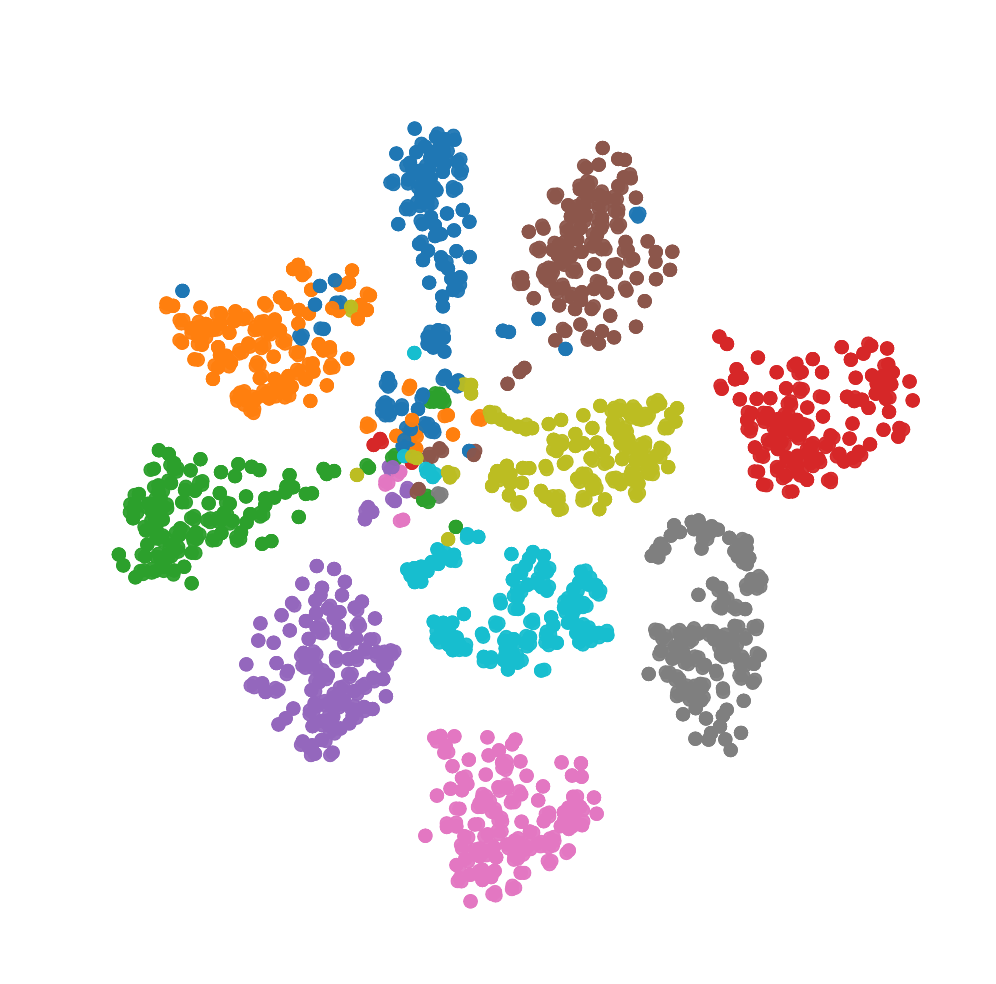}
     \end{minipage}}
  \subfigure[Trunk block]{
    \begin{minipage}[b]{.3\linewidth}
        \centering
        \includegraphics[height=3cm]{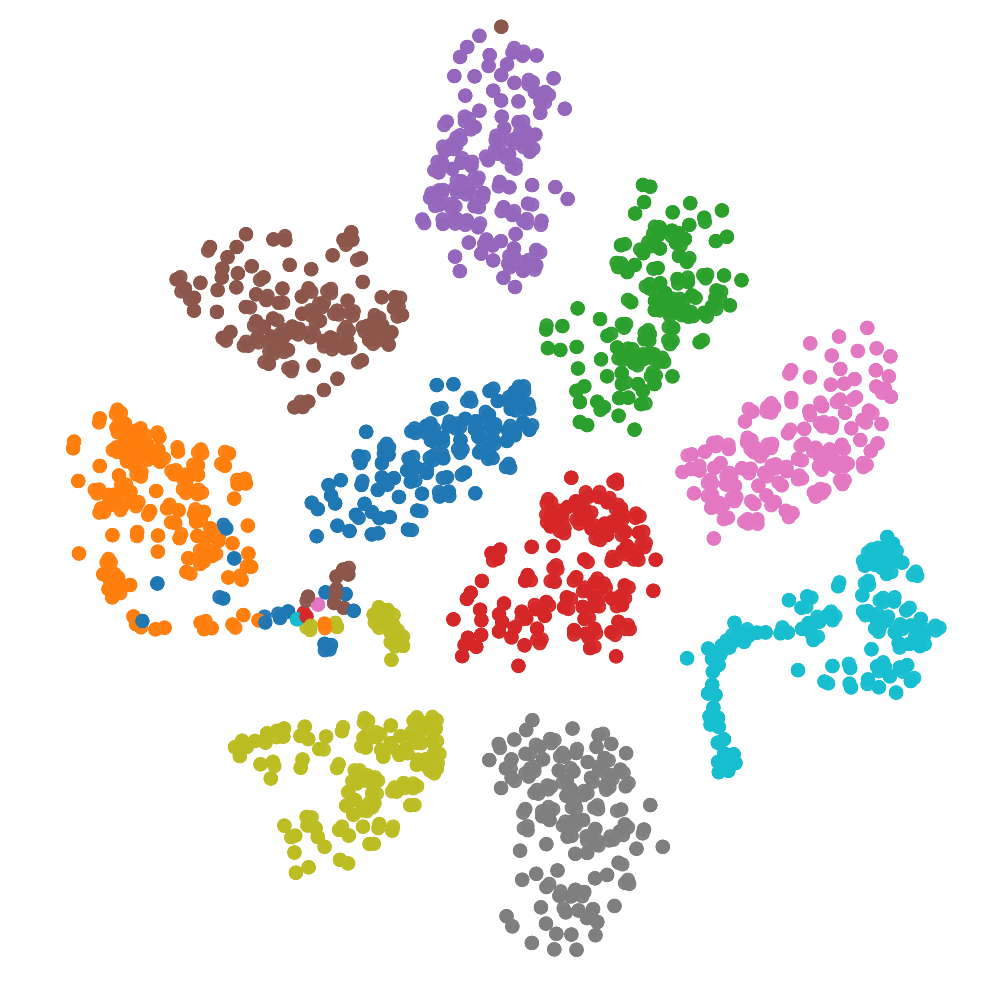}
    \end{minipage}}
  \subfigure[TBCNet]{
    \begin{minipage}[b]{.3\linewidth}
        \centering
        \includegraphics[height=3cm]{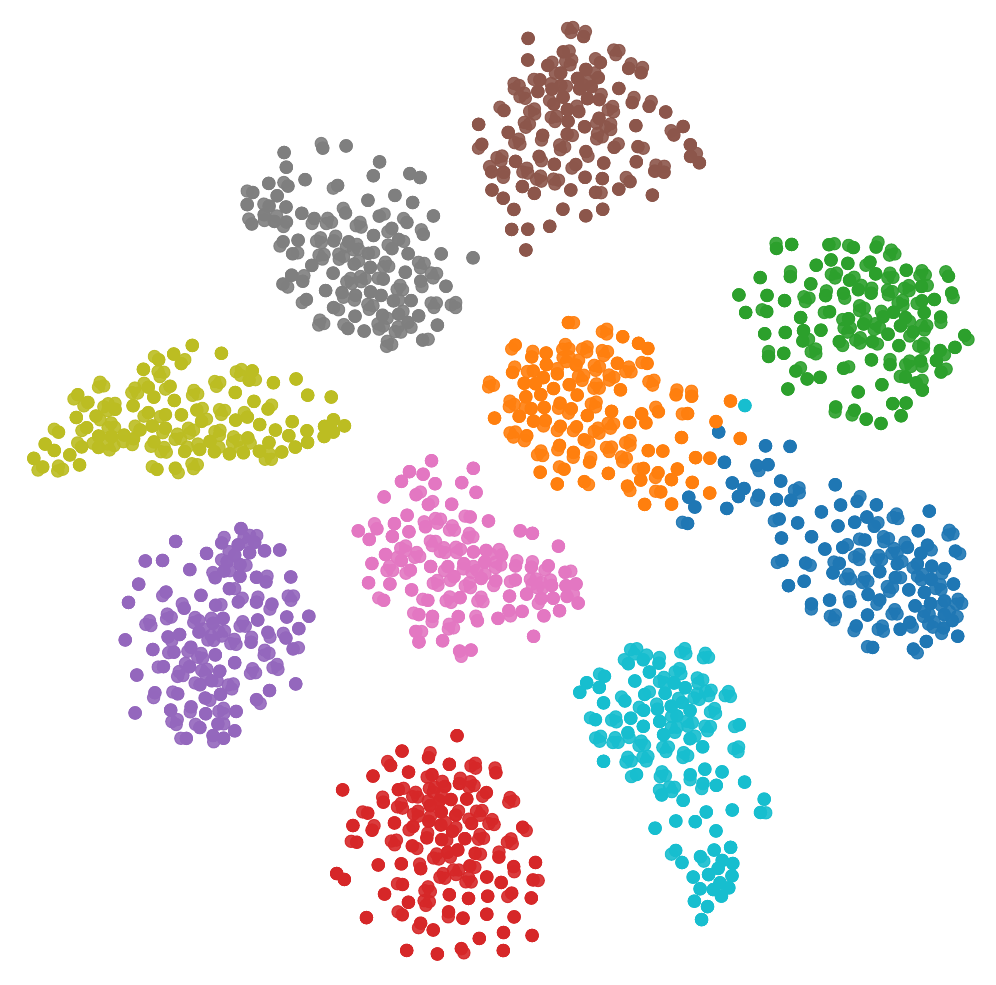}
    \end{minipage}}
  \caption{The t-SNE visualizations of the output derived from three models. Different colored points represent the first ten classes of NTU-RGB+D 60 on the CS benchmark.}
  \label{img10}
\end{figure}

Fig. \ref{img10} shows t-SNE embedding visualizations of feature distributions in three distinct models: ResNet3D, Trunk block, and TBCNet. Unlike the class mixing observed in Fig. \ref{img10}(a), the trunk block effectively segregates features from different classes, as depicted in Fig. \ref{img10}(b).
However, several classes exhibit partial entanglement like ‘throw' (light blue class) and ‘brushing teeth’ (golden class).
Fig. \ref{img10}(c) presents results in TBCNet, demonstrating superior clustering quality and the clearer boundaries. For instance, ‘drinking water’ (blue class) is more distinctly separated from ‘eating/snacking’ (orange class) and ‘brushing teeth’ (golden class). It suggests efficacy of our method in distinguishing similar actions.

\subsection{Limitation} \label{sec4.6}
Based on the overall analysis of the above experiments, we summarized the limitations and areas for improvement as follows:

\textbf{Two-stage training.} 
The proposed approach employs a two-stage training strategy. The contrastive learning is integrated after the preliminary training of both the trunk block and branch block. This ensures effective information distillation but may increase computational costs.

\textbf{In-adaptability to single-view scenarios.} 
MVDA used in the trunk block is designed primarily to integrate multi-view spatial information and cross-view association, its performance is restricted in single-view scenarios.

\textbf{Enhancements to branch block.} 
Given the branch block can be removed during inference, future research will focus on more advanced designs and incorporate other modalities to enhance overall performance.

%------------------------------------------------------------------------- 
\section{Conclusion}
In this work, we present a novel trunk-branch contrastive network (TBCNet) for RGB-based multi-view action recognition. 
Different from previous architectures, the network facilitates aggregated multi-view features to adequately absorb distinct information from each view via the proposed trunk-branch contrastive learning. And a multi-view deformable aggregation (MVDA) utilizes the global aggregation module (GAM) and deformable attention with composite relative position bias (CRPB) to realize multi-view fusion and cross-view correlations. Notably, the branch block which provides vital details can be discarded during inference, reducing actual computational costs.
Compared with other RGB-based works, the method exhibits SOTA performance under cross-subject conditions on various multi-view datasets. 
Future research will incorporate skeleton information into the branch block to leverage multi-modal fusion to boost model performance, mitigating reliance on skeleton data and maintaining efficient inference in actual scenes.

\bibliographystyle{unsrt}  


% George Kour and Raid Saabne.
% \newblock Real-time segmentation of on-line handwritten arabic script.
% \newblock In {\em Frontiers in Handwriting Recognition (ICFHR), 2014 14th
%   International Conference on}, pages 417--422. IEEE, 2014.

% George Kour and Raid Saabne.
% \newblock Fast classification of handwritten on-line arabic characters.
% \newblock In {\em Soft Computing and Pattern Recognition (SoCPaR), 2014 6th
%   International Conference of}, pages 312--318. IEEE, 2014.

% \bibitem{hadash2018estimate}
% Guy Hadash, Einat Kermany, Boaz Carmeli, Ofer Lavi, George Kour, and Alon
% \newblock Estimate and replace: A novel approach to integrating deep neural
%   networks with existing applications.
% \newblock {\em arXiv preprint arXiv:1804.09028}, 2018.

% \end{thebibliography}
\bibliography{ref}   %ref为.bib文件名

\end{document}